\documentclass[10pt,a4paper]{article}

\usepackage{iftex}
\ifPDFTeX
  \usepackage[utf8]{inputenc}
\fi
\usepackage[T1]{fontenc}
\usepackage{lmodern}

\usepackage{microtype}

\newcommand{\AurekaDisplayFont}{\sffamily\bfseries}
\newcommand{\AurekaUtilityFont}{\sffamily\mdseries\footnotesize\textls[80]}
\usepackage[a4paper,top=27mm,bottom=25mm,left=31mm,right=31mm,headheight=30pt,headsep=14pt]{geometry}
\usepackage{graphicx}
\usepackage{subcaption}
\usepackage{svg}
\usepackage{xcolor}
\usepackage{amsmath,amssymb,amsthm,mathtools}
\usepackage{booktabs}
\usepackage{array}
\usepackage{enumitem}
\usepackage{caption}
\usepackage{subcaption}
\usepackage{tikz}
\usepackage{eso-pic}
\usepackage[most]{tcolorbox}
\usepackage{titlesec}
\usepackage{fancyhdr}
\usepackage{authblk}
\usepackage[numbers,sort&compress]{natbib}
\usepackage[colorlinks=true,linkcolor=AurekaBlue,citecolor=AurekaBlue,urlcolor=AurekaBlue]{hyperref}
\usepackage[nameinlink,noabbrev]{cleveref}
\usepackage{indentfirst}
\usepackage[bottom]{footmisc}

\usepackage{xcolor}
\usepackage{xurl}

\usepackage{fontawesome5}

\definecolor{linkpink}{RGB}{255,0,140}

\newcommand{\pinkurl}[1]{%
  {\hypersetup{urlcolor=linkpink}\url{#1}}%
}

\usepackage[most]{tcolorbox}
\usepackage{xcolor}
\usepackage{enumitem}

\definecolor{contribFrame}{HTML}{9B93F5}
\definecolor{contribTitle}{HTML}{A9A2F7}
\definecolor{contribBack}{HTML}{F5F4FF}

\definecolor{AurekaInk}{HTML}{1B1E28}
\definecolor{AurekaMuted}{HTML}{5F6675}
\definecolor{AurekaLine}{HTML}{E6E8EF}
\definecolor{AurekaSoft}{HTML}{FAF8FF}
\definecolor{AurekaPurple}{HTML}{8F7BFF}
\definecolor{AurekaIndigo}{HTML}{6A5DBA}
\definecolor{AurekaPurpleLight}{HTML}{DCD5FF}
\definecolor{AurekaIndigoLight}{HTML}{BEB7EA}
\definecolor{AurekaTopbar}{HTML}{B8AAFF}
\definecolor{AurekaBlue}{HTML}{2454A6}
\definecolor{AurekaSeedRed}{HTML}{E84B4B}

\newcommand{\AurekaLogoFile}{aureka_logo.pdf}
\newcommand{\AurekaLogo}{\includegraphics[height=8mm]{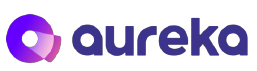}}

\newcommand{\FirstPageBrandAccent}{%
  \AddToShipoutPictureBG*{%
    \begin{tikzpicture}[remember picture,overlay]
      \fill[AurekaTopbar] (current page.north west) rectangle ([yshift=-1.2mm]current page.north east);
      \fill[AurekaSeedRed] ([xshift=0.68\paperwidth]current page.north west) rectangle ([xshift=0.78\paperwidth,yshift=-1.2mm]current page.north west);
    \end{tikzpicture}%
  }%
}

\fancypagestyle{plain}{%
  \fancyhf{}%
  \fancyhead[L]{\AurekaLogo}%
  \fancyhead[R]{\AurekaUtilityFont{\textcolor{AurekaMuted}{AUREKA AI RESEARCH}}}%
  \fancyfoot[C]{\AurekaUtilityFont{\textcolor{AurekaMuted}{\thepage}}}%
}

\titleformat{\section}
  {\Large\AurekaDisplayFont\color{AurekaInk}}
  {\thesection}{0.65em}{}
  [{\vspace{-0.10em}\color{AurekaPurple}\titlerule[1.0pt]}]

\titleformat{\subsection}
  {\large\AurekaDisplayFont\color{AurekaIndigo}}
  {\thesubsection}{0.65em}{}

\titleformat{\subsubsection}
  {\normalsize\AurekaDisplayFont\color{AurekaIndigo}}
  {\thesubsubsection}{0.65em}{}

\titlespacing*{\section}{0pt}{2.2em plus 0.4em minus 0.2em}{0.95em}
\titlespacing*{\subsection}{0pt}{1.45em plus 0.25em minus 0.15em}{0.55em}
\titlespacing*{\subsubsection}{0pt}{1.1em plus 0.2em minus 0.1em}{0.4em}

\setlist[itemize]{leftmargin=1.4em,itemsep=0.25em,topsep=0.25em}
\setlist[enumerate]{leftmargin=1.6em,itemsep=0.25em,topsep=0.25em}
\newtcolorbox{insightbox}[1][]{
  enhanced,
  colback=AurekaSoft,
  colframe=AurekaPurpleLight,
  boxrule=0.6pt,
  arc=2mm,
  left=1.1em,
  right=1.1em,
  top=0.85em,
  bottom=0.85em,
  title=#1,
  fonttitle=\sffamily\bfseries,
  coltitle=AurekaInk
}

\newtcolorbox{methodbox}[1][]{
  enhanced,
  colback=white,
  colframe=AurekaLine,
  boxrule=0.6pt,
  arc=2mm,
  left=1.1em,
  right=1.1em,
  top=0.85em,
  bottom=0.85em,
  borderline west={2pt}{0pt}{AurekaIndigoLight},
  title=#1,
  fonttitle=\sffamily\bfseries,
  coltitle=AurekaInk
}

\theoremstyle{definition}

\theoremstyle{plain}

\title{\vspace{-0.4em}\AurekaDisplayFont\color{AurekaInk}\LARGE
Folding, Reasoning, and Scaling \\ with Open-source Drug Discovery Engine}
\author{\small OpenDDE Project, Aureka AI Research \thanks{Full author list in Contributions}}
\date{}

\usepackage{setspace}
\begin{document}
\thispagestyle{fancy}

\FirstPageBrandAccent

\maketitle
\vspace{-1.5em}

\begin{insightbox}[Abstract]

Accurately modeling biomolecular interactions is a central bottleneck in biology and therapeutic discovery. Here, we introduce Open Drug Discovery Engine (\textsc{OpenDDE}), an open-source, all-atom biomolecular foundation model that uses co-folding as the entry point to a scalable AI-driven drug discovery engine. Rather than treating structure prediction as an isolated endpoint, \textsc{OpenDDE} is designed as a shared structural reasoning layer for modeling sequence–structure–function relationships across biomolecular complexes, enabling complex structure prediction today while providing a foundation for \textit{de novo} design, affinity estimation, structure-conditioned optimization, and more. 
\textsc{OpenDDE} integrates advances in all-atom architecture, atomic latent reasoning, inference optimization, and large-scale data processing to achieve IsoDDE-level co-folding accuracy within a reproducible and openly accessible framework. We also identify two scaling-law directions for co-folding models, revealing practical routes for continued improvement through data, model, inference, and training scaling. By releasing training code, inference pipelines, checkpoints, and benchmarks, \textsc{OpenDDE} aims to democratize access to frontier biomolecular intelligence, accelerate global collaboration, and lay an open foundation for next-generation drug discovery systems that can move from predicting molecular structures toward designing, scoring, and optimizing therapeutic candidates for human health.\\


\vspace{-0.3cm}
\noindent\faGithub\ \textbf{GitHub:} \pinkurl{https://github.com/aurekaresearch/OpenDDE}\\

\vspace{-0.3cm}
\noindent\faDatabase\ \textbf{HuggingFace:} \pinkurl{https://huggingface.co/aurekaresearch/OpenDDE}\\


\end{insightbox}

\begin{figure}
\begin{subfigure}[!t]{\linewidth}
    \centering
    \includegraphics[width=0.98\linewidth]{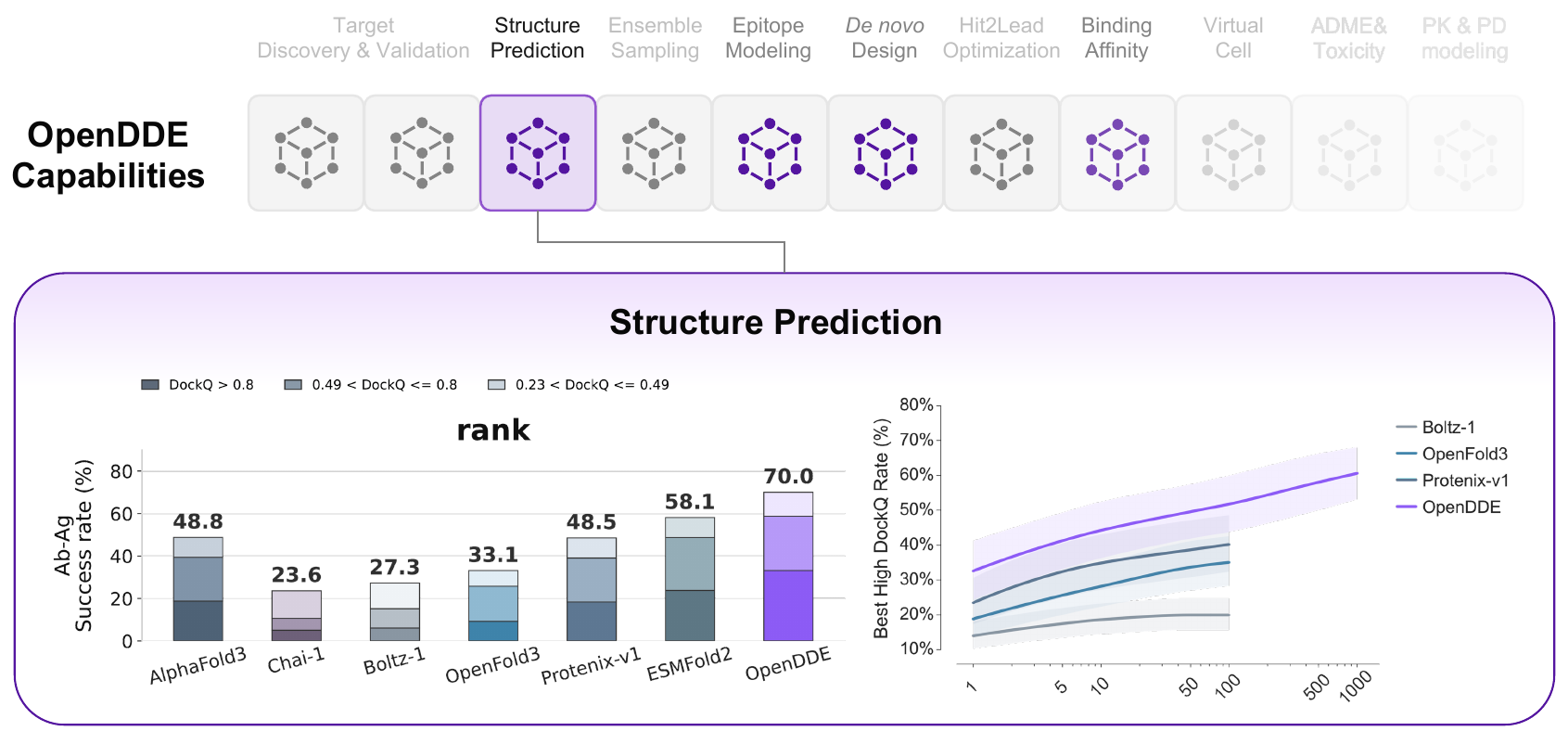}
\end{subfigure}
\end{figure}

\section*{Overview}

\begin{tcolorbox}[
    enhanced,
    breakable,
    colback=contribBack,
    colframe=contribFrame,
    colbacktitle=contribTitle,
    coltitle=black,
    title={Key Contributions},
    fonttitle=\bfseries\large,
    boxrule=1.2pt,
    arc=3mm,
    left=12pt,
    right=12pt,
    top=10pt,
    bottom=10pt,
    toptitle=5pt,
    bottomtitle=5pt
]

\begin{enumerate}[leftmargin=*, itemsep=0.8em]

    \item \textbf{Atomic latent reasoning over biomolecular tokens.}
    We incorporate latent reasoning into biomolecular modelling, enabling the model to refine representations of local geometry, chemical context, and cross-molecular interfaces before all-atom structure generation.

    \item \textbf{A folding-centered foundation for an extensible drug-discovery engine.}
    \textsc{OpenDDE} currently focuses on structure prediction, but its unified architecture is designed to support future \textit{de novo} molecular design, affinity prediction, and other structure-conditioned modules.

    \item \textbf{Scaling laws and data distillation.}
    We study scaling directions along model-parameter and data axes, and revise distillation strategies for monomeric and multimeric structures to improve robustness, generalization, and training efficiency.
\end{enumerate}

\end{tcolorbox}

Characterising biomolecular interactions among proteins, nucleic acids, small-molecule ligands and other cellular components is fundamental to understanding biological mechanisms and to modulating disease-relevant processes through therapeutic intervention \cite{kuntz1992structure, santos2017comprehensive}. In silico models that can predict these interactions with experimental-level accuracy would therefore provide a structural basis for scalable and reliable drug discovery.

AlphaFold2 \cite{varadi2022alphafold} transformed protein structure prediction by achieving near-experimental accuracy for many protein monomers.  AlphaFold3 \cite{abramson2024accurate} extended this paradigm to joint, atomic-level modelling of complexes containing proteins, nucleic acids, small molecules, ions and modified residues. More broadly, AlphaFold-style and structure-conditioned models have become reusable engines for epitope prediction, binder design \cite{watson2023novo, stark2025boltzgen, zhang2025odesign}, virtual screening \cite{lu2022tankbind, passaro2025boltz}, conformational sampling \cite{lu2024dynamicbind, lewis2025scalable} and the acceleration or interpretation of experimental workflows \cite{pacesa2025one}. The recent introduction of IsoDDE further shows that co-folding can serve as the core of an AI drug discovery engine, with reported gains in protein-ligand generalisation, antibody-antigen interface prediction, pocket identification and binding affinity estimation. However, the most capable system remains closed, limiting reproducibility, independent validation and community-driven extension. At the same time, incremental improvements in co-folding accuracy have begun to slow \cite{vskrinjar2025have}, raising a central question for the field: \textit{whether current tokenization, data curation, model architectures, and inference procedures are approaching practical bottlenecks, and where the next scaling directions should come from?}

In this manuscript, we present OpenDDE, to our knowledge the first Apache-licensed all-atom generative foundation model reported to reach IsoDDE-level co-folding accuracy. Building on recent open co-folding systems such as Protenix-v1 and OpenFold3, \textsc{OpenDDE} introduces three advances that extend the co-folding paradigm across both pre-training and post-training:






In the following sections, we present the model architecture, training, inference, and scaling laws behind \textsc{OpenDDE}, followed by benchmarks against open and closed folding systems across diverse biomolecular complex prediction tasks. \textsc{OpenDDE} performs on par with state-of-the-art closed models and substantially improves over existing models on antibody-antigen generalization benchmarks.

As an evolving project, \textsc{OpenDDE} and its repository will be regularly updated with contributions from our team and the broader community. We release checkpoints, code, inference pipelines, and benchmarks under the Apache-2.0 license to enable the community to explore, refine, and deploy biological intelligence at scale. Beyond strong co-folding performance, \textsc{OpenDDE} provides an open platform for studying how folding, reasoning, and scaling can advance biomolecular intelligence.



\section{Performance}

\subsection{Structure Prediction Result}

\begin{figure}[ht!]
\centering
    \includegraphics[width=\linewidth]{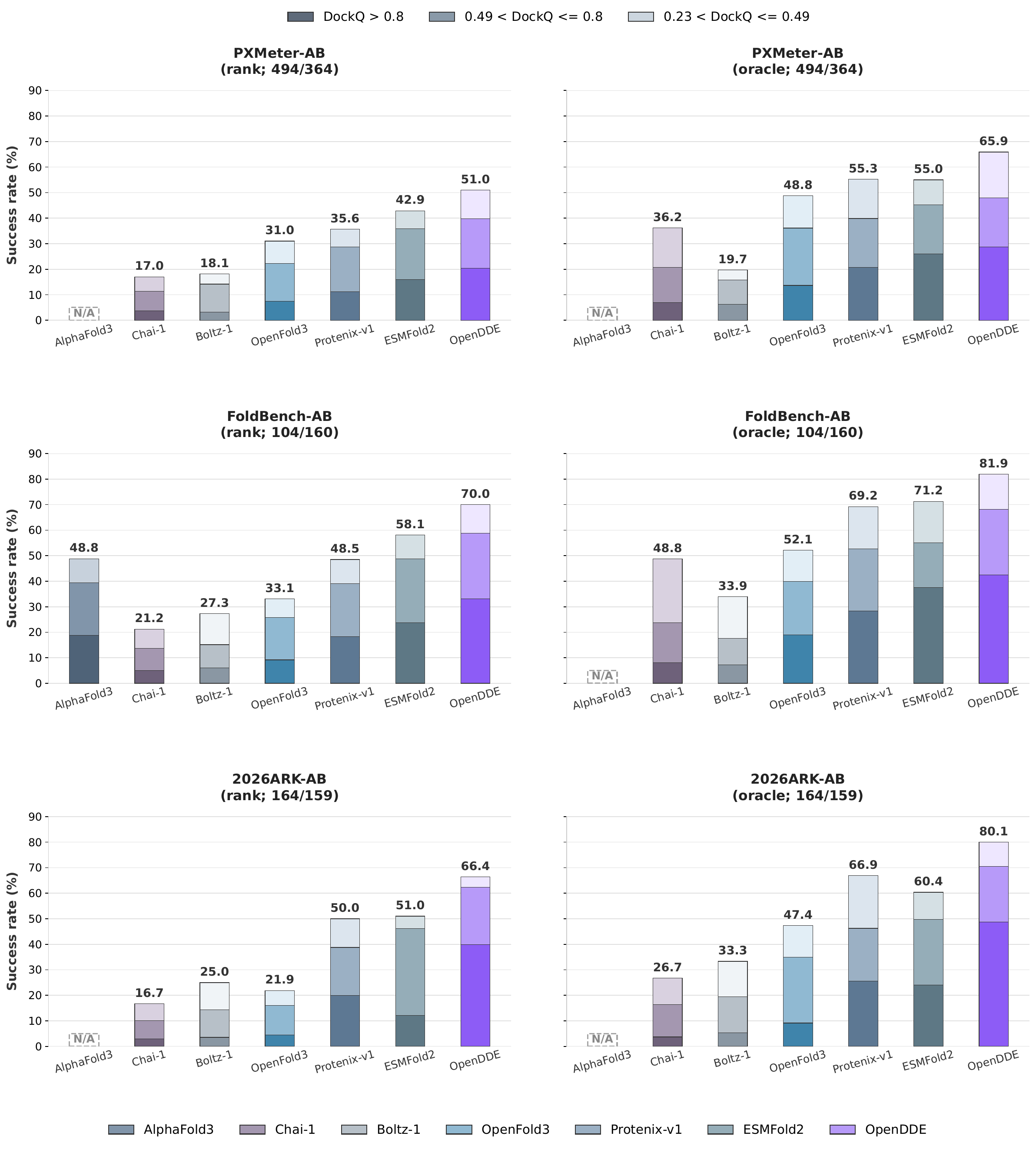}
    \caption{\textbf{Antibody--antigen structure prediction performance across PXMeter-AB, FoldBench-AB, and 2026ARK-AB.} Results are reported as DockQ success rates, decomposed into acceptable-, medium-, and high-quality predictions. \textit{Left:} Ranking-based selection. \textit{Right:} Oracle-based selection.}
\label{fig:abag_benchmark}
\end{figure}






\textbf{Benchmarking \textsc{OpenDDE} on Antibody-Antigen Dataset (Figure~\ref{fig:abag_benchmark}).}
Antibody-antigen complexes provide a stringent test for structure-based discovery, as their therapeutic relevance is matched by the difficulty of accurately modelling flexible and chemically diverse binding interfaces. Figure~\ref{fig:abag_benchmark} compares \textsc{OpenDDE} with AlphaFold3 \cite{abramson2024accurate}, Chai-1 \cite{chai2024chai}, Boltz-1 \cite{wohlwend2025boltz}, OpenFold3 \cite{openfold3-preview}, Protenix-v1 \cite{bytedance2025protenix}, and ESMFold2 \cite{candido2026language} across three antibody-antigen benchmark collections. Success rates are decomposed by interface quality using DockQ thresholds, separating acceptable, medium-quality, and high-quality predictions. \textsc{OpenDDE} achieves the highest overall success rate on all evaluated benchmarks, reaching 51.0\% on PXMeter-AB, 70.0\% on FoldBench-AB, and 66.4\% on the 2026ARK-AB benchmark. 

These results substantially improve over the strongest baselines, including ESMFold2 on PXMeter-AB and FoldBench-AB and ESMFold2/Protenix-v1 on 2026ARK-AB. Importantly, the gains are not limited to low-threshold recovery: \textsc{OpenDDE} consistently increases the fraction of medium- and high-DockQ interfaces, indicating improved modeling of antibody-antigen geometry rather than merely producing marginally acceptable complexes. Together, these results show that \textsc{OpenDDE} provides a strong open-source foundation for antibody-antigen co-folding and brings a broader set of therapeutic targets into a practically useful accuracy regime.


\textbf{Structure Prediction of New Biological Systems (Figure~\ref{fig:abag_benchmark}).}
We examine \textsc{OpenDDE} on newly released antibody-antigen structures by curating the 2026ARK-AB benchmark, following the data protocol of PXMeter \cite{ma2025dataset}. This benchmark contains 164 PDB complexes, corresponding to 159 unique antibody-antigen interface clusters, and is curated to assess prediction performance on recent, low-homology systems that are not available to earlier released models. Antibody-antigen interfaces are defined by residue-level spatial proximity and annotated using SAbDab \cite{dunbar2014sabdab}. We retain high-quality biological assemblies using standard filters for structure quality, assembly size, chain completeness, and interface geometry. Protein entities are clustered with MMseqs2 easy-cluster using a 40\% minimum sequence identity threshold and an 80\% alignment coverage requirement, and each interface cluster is defined as the unordered pair of the two interacting entity clusters. 

On this newest benchmark, \textsc{OpenDDE} achieves the highest overall success rate, correctly predicting 66.4\% of interfaces at DockQ $>$ 0.23. It outperforms ESMFold2 (51.0\%), Protenix-v1 (50.0\%), OpenFold3 (21.9\%), Boltz-1 (25.0\%), and Chai-1 (16.7\%). \textsc{OpenDDE} also shows a clear advantage in the high-fidelity regime, with a large fraction of complexes reaching DockQ $>$ 0.8. These results demonstrate \textsc{OpenDDE}'s strong ability to predict newly released antibody-antigen interfaces and suggest improved generalization to emerging biological systems.

\begin{figure}[t!]
\centering
    \includegraphics[width=\linewidth]{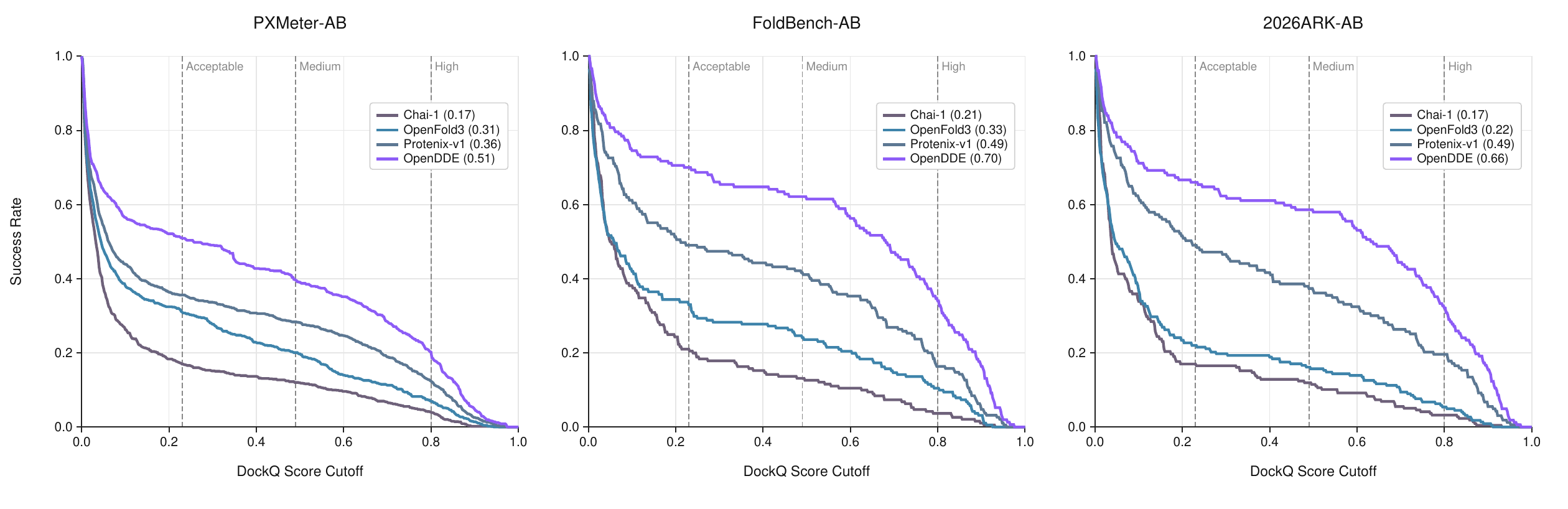}
    \caption{\textbf{Cumulative distribution of DockQ across PXMeter-AB, FoldBench-AB, and 2026ARK-AB.} Each curve shows the fraction of targets whose top-ranked prediction reaches DockQ $\ge$ the x-axis cutoff, for Chai-1, OpenFold3, Protenix-v1, and OpenDDE; curves closer to the upper-right are better.}
\label{fig:survive}
\end{figure}

\textbf{New Model Capacity (Figure~\ref{fig:abag_benchmark}).}
We evaluate model capacity under an oracle selection setting, where the best prediction is selected from each model's generated candidates. This setting approximates the maximal achievable performance of a model when ranking or confidence estimation is not the limiting factor, and therefore reflects the upper bound of its sampling capacity. As shown in Figure~\ref{fig:abag_benchmark}, \textsc{OpenDDE} achieves the highest oracle-selected success rate across all antibody-antigen benchmarks, reaching 65.9\% on PXMeter-AB, 81.9\% on FoldBench-AB, and 80.1\% on 2026ARK-AB. These results substantially exceed the corresponding non-oracle benchmark performance, indicating that \textsc{OpenDDE} frequently generates high-quality antibody-antigen conformations even when the top-ranked prediction is not always optimal. The large oracle gains suggest that \textsc{OpenDDE} has strong latent modeling capacity for antibody-antigen interfaces, and that further improvements in confidence estimation or candidate selection could translate this capacity into higher realized prediction accuracy.

\textbf{DockQ Success-rate Curve (Figure~\ref{fig:survive}).} 
The DockQ success-rate curves provide a threshold-sweep view of interface prediction quality. For each DockQ cutoff, the curve reports the fraction of targets whose predicted interface exceeds that quality threshold, so curves that remain higher across the x-axis indicate better performance over the full range from acceptable to high-quality predictions. Across PXMeter-AB, FoldBench-AB, and 2026ARK-AB, \textsc{OpenDDE} consistently maintains the highest curve, showing that its advantage is not restricted to a single DockQ threshold. The gap is especially clear in the medium- and high-quality regimes, where \textsc{OpenDDE} retains a larger fraction of successful predictions as the cutoff becomes stricter. This indicates that \textsc{OpenDDE} improves both coarse interface recovery and high-fidelity antibody-antigen docking accuracy, producing more predictions with accurate binding geometry rather than only increasing low-threshold success.

\subsection{Scaling Laws in Biomolecular Foundation Models}

\begin{figure}[ht!]
\centering
    \includegraphics[width=\linewidth]{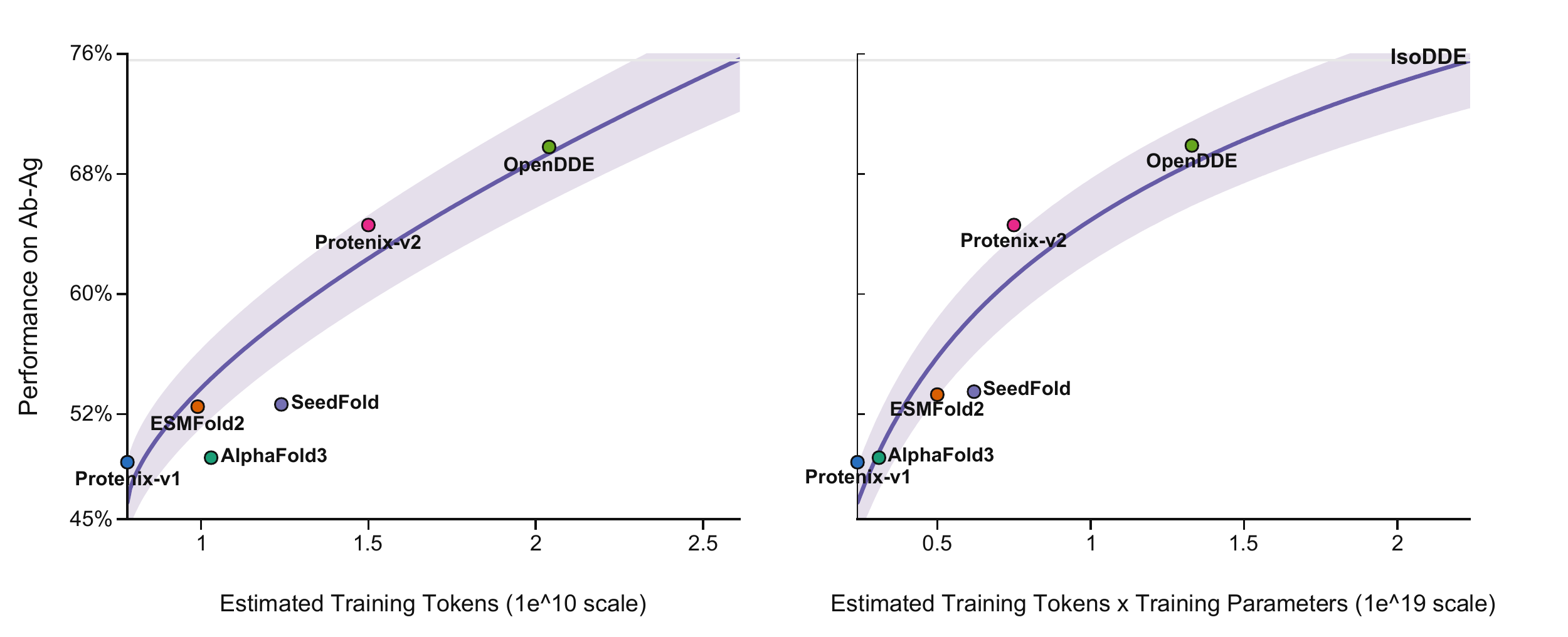}
    \caption{\textbf{Scaling laws in biomolecular foundation models.} \textit{Left:} Model performance as a function of estimated training tokens, computed as training steps $\times$ number of GPUs (batch size) $\times$ crop tokens. \textit{Right:} Model performance as a function of estimated training cost, computed as estimated training tokens $\times$ trainable model parameters. }
\label{fig:abag_scaling}
\end{figure}

We observe a clear scaling trend in biomolecular foundation models. To examine this trend, we compare \textsc{OpenDDE} with AlphaFold3 \cite{abramson2024accurate}, Protenix-v1 \cite{bytedance2025protenix}, Protenix-v2 \cite{zhang2026protenix}, SeedFold \cite{zhou2025seedfold}, and ESMFold2 \cite{candido2026language} on antibody--antigen structure prediction (Figure~\ref{fig:abag_scaling}). We use two estimates of training scale. The first is the estimated number of training tokens, computed as training steps $\times$ number of GPUs (batch size) $\times$ crop tokens.\footnote{For OpenDDE, this is estimated as $400\times(10\mathrm{k}\times384+60\mathrm{k}\times384+15\mathrm{k}\times544+15\mathrm{k}\times544+10\mathrm{k}\times768)\approx2.04\times10^{10}$.} The second is the estimated training cost, computed as estimated training tokens $\times$ trainable model parameters.\footnote{For OpenDDE, this is estimated as $2.04\times10^{10}\times655\mathrm{M}\approx1.33\times10^{19}$.}

\textbf{Evidence of Scaling Laws (Figure~\ref{fig:abag_scaling}).}
The scaling-law curves show a clear positive relationship between training scale and Ab-Ag performance. As the estimated training compute increases, model performance improves from the lower-scale models such as Protenix-v1 and AlphaFold3 toward higher-performing models such as Protenix-v2 and OpenDDE. The trend is smooth and sublinear, resembling scaling behavior observed in large language models: early increases in scale produce relatively large gains, while later gains become more gradual as the curve approaches the IsoDDE reference level.

\textsc{OpenDDE} sits at the high-scale, high-performance end of the curve, achieving the best observed Ab-Ag performance among the compared models and narrowing the gap to IsoDDE. The remaining gap suggests that further scaling may continue to improve performance, but with diminishing returns. Deviations among individual models also indicate that scale alone does not fully determine performance; architecture, data quality, and training strategy likely contribute to where each model falls relative to the overall scaling trend.

\textbf{Entering the Era of Scaling (Figure~\ref{fig:abag_scaling}).} Across these models and performance, the results improve with both data scale and effective compute scale. This observation suggests that biomolecular structure modeling is entering the same scaling regime that has transformed (large) language models: \textit{larger effective training corpora, larger-scale models, and greater compute can systematically translate into stronger biological reasoning and structure prediction capability}.

\subsection{Test-Time Scaling}

\begin{figure}[ht!]
\centering
    \includegraphics[width=\linewidth]{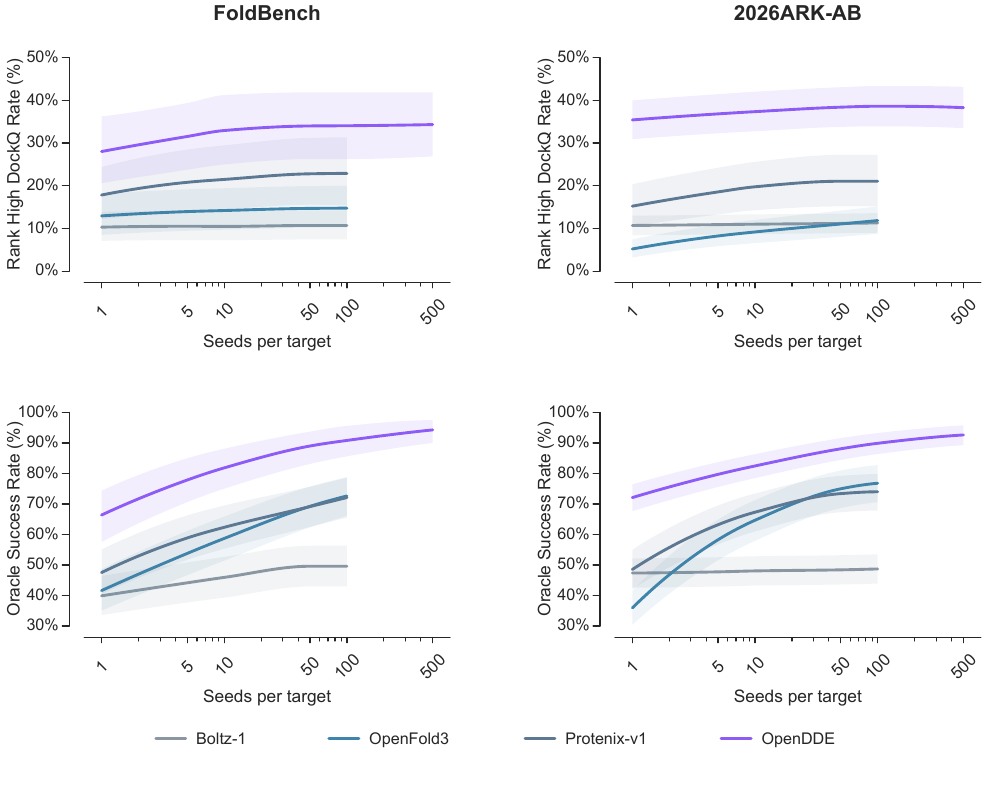}
    \vspace{-0.8cm}
    \caption{\textbf{Test-time scaling on FoldBench-AB and 2026ARK-AB.} \textit{Top:} Ranking-based selection of DockQ $>$ 0.8. \textit{Bottom}: Oracle-based selection.}
\label{fig:tts}
\end{figure}






We evaluate whether additional inference-time sampling improves antibody-antigen prediction performance (Figure~\ref{fig:tts}). For each target, we generate multiple stochastic samples with different random seeds and select the final prediction using the model ranking score. We report two settings: ranking-based high-quality success rate with DockQ $>$ 0.8, and oracle success rate, where the best candidate among all generated samples is selected using the ground-truth DockQ. The oracle setting measures the maximal performance available in the sampled candidate set and separates sampling capacity from ranking quality.

As shown in Figure~\ref{fig:tts}, \textsc{OpenDDE} exhibits consistent test-time scaling on both FoldBench-AB and 2026ARK-AB. On FoldBench-AB, increasing the number of seeds improves the ranked high-DockQ rate from roughly 28\% with one seed to about 34\% at large sampling budgets. Similar behavior is observed on 2026ARK-AB, where \textsc{OpenDDE} improves from approximately 35\% to 38\% ranked high-DockQ successThese gains indicate that additional samples expose better antibody-antigen conformations rather than simply producing redundant predictions.

The oracle curves show a larger scaling effect. On FoldBench-AB, \textsc{OpenDDE}'s oracle success rises from about 66\% with one seed to above 90\% with hundreds of seeds. On 2026ARK-AB, the oracle success similarly increases from about 67\% to nearly 90\%. This gap between ranked and oracle performance suggests that \textsc{OpenDDE} often generates high-quality antibody-antigen structures that are not always selected as the top-ranked prediction. Therefore, test-time scaling reveals strong latent sampling capacity, and further improvements in confidence calibration or candidate ranking could convert more of this oracle performance into realized prediction accuracy.

Across both metrics, \textsc{OpenDDE} scales more favorably with inference-time compute than the compared open models, including Boltz-1, OpenFold3, and Protenix-v1. These results show that \textsc{OpenDDE} benefits predictably from larger sampling budgets and that its diffusion-based generation process provides a practical route for improving antibody-antigen prediction accuracy at test time.

\subsection{Case Study}
\begin{figure}[t!]
\centering
    \includegraphics[width=\linewidth]{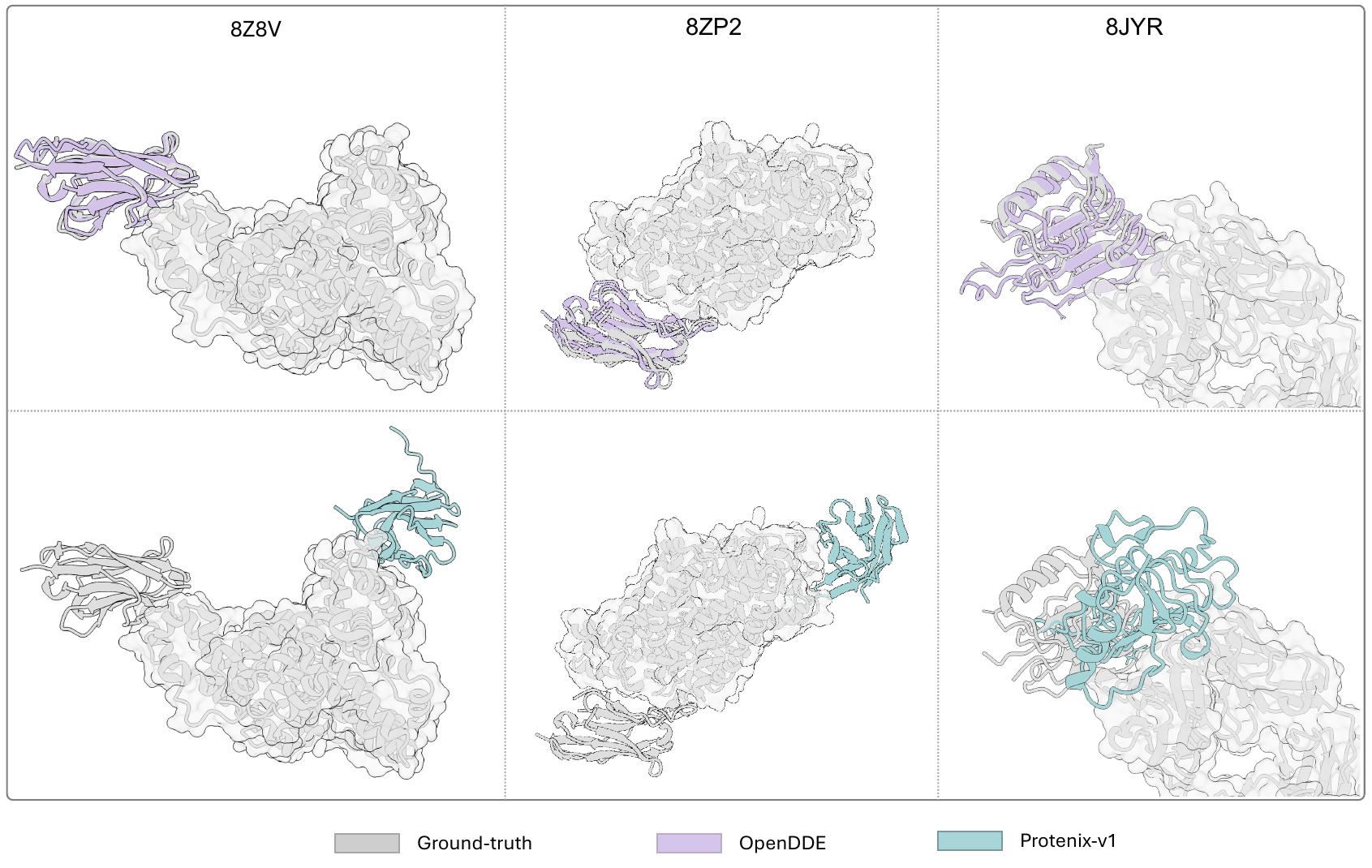}
    \caption{\textbf{Case studies of antibody-antigen structure predictions, from Protenix-v1 and OpenDDE.} \textit{Left:} ALB8 VHH binding to human serum albumin (PDB: 8Z8V, 2024-09). \textit{Middle:} Antibody binding to human norepinephrine transporter NET (PDB: 8ZP2, 2024-09). \textit{Right:} Anti-HER2 Fab H2Mab-119 binding to HER2 domain I (PDB: 8JYR, 2024-03). OpenDDE predictions are shown in purple; Protenix-v1 predictions are shown in teal, each superimposed on the ground-truth structure are shown in gray.}
\label{fig:case.study}
\end{figure}

We show predicted structures of new biological systems in Figure~\ref{fig:case.study}. OpenDDE predictions, colored in purple, closely recover the experimentally resolved binding geometry shown in gray across all three systems. In 8Z8V, OpenDDE places the ALB8 VHH on the correct surface of human serum albumin, preserving both the global orientation and the local interface geometry. In 8ZP2, OpenDDE recovers the nanobody position on the NET transporter complex, despite the added difficulty of a membrane protein target and cryo-EM-derived structure. In 8JYR, OpenDDE accurately aligns the antibody variable region against HER2 domain I, capturing the correct epitope-facing pose.

These case studies suggest that OpenDDE improves structure prediction at the level most relevant for biological discovery: the geometry of molecular recognition. While the examples are qualitative and should be complemented by cutoff-controlled quantitative benchmarks, their recent release dates make them useful stress tests for assessing OpenDDE on newly solved biological systems rather than only on saturated historical cases.

\section{Training Details}

\subsection{Model Parameters}
\textsc{OpenDDE} has 655M training parameters using NVIDIA computing kernels. We report model, training, and sampling hyperparameters in Table~\ref{tab:opendde_hyperparameters}.

\begin{table}[ht!]
\centering
\caption{Model, training, and sampling hyperparameters of OpenDDE.}
\label{tab:opendde_hyperparameters}
\small
\begin{minipage}[t]{0.44\linewidth}
\centering
\caption*{(a) Model hyperparameters}
\resizebox{\linewidth}{!}{
\begin{tabular}{lc}
\hline
\textbf{Parameter} & \textbf{Value} \\
\hline
$c_s$ & 384 \\
$c_z$ & 384 \\
$c_{\mathrm{token}}$ & 384 \\
$c_{\mathrm{atom}}$ & 128 \\
$c_{\mathrm{atompair}}$ & 16 \\
$c_{s,\mathrm{inputs}}$ & 449 \\
Pairformer blocks & 48 \\
MSA module blocks & 4 \\
MSA channel $c_m$ & 128 \\
Diffusion transformer blocks & 24 \\
Diffusion atom encoder blocks & 3 \\
Diffusion atom decoder blocks & 3 \\
Structural token roles & 7 \\
Structural token refiner & 4 blocks, 8 heads \\
Distogram bins & 96 \\
Distogram range & [2.25, 25.75] \\
\hline
\end{tabular}
}
\end{minipage}
\hfill
\begin{minipage}[t]{0.48\linewidth}
\centering
\caption*{(b) Training and sampling hyperparameters}
\resizebox{\linewidth}{!}{
\begin{tabular}{lc}
\hline
\textbf{Parameter} & \textbf{Value} \\
\hline
\texttt{model.N\_cycle} & 4 \\
\texttt{sample\_diffusion.N\_step} & 20 \\
\texttt{sample\_diffusion.N\_sample} & 5 \\
\texttt{sample\_diffusion.N\_step\_mini\_rollout} & 20 \\
\texttt{sample\_diffusion.N\_sample\_mini\_rollout} & 1 \\
\texttt{diffusion\_batch\_size} & 48 \\
\texttt{diffusion\_chunk\_size} & 4 \\
\texttt{train\_chunk\_size} & \texttt{None} \\
\texttt{blocks\_per\_ckpt} & 1 \\
\texttt{train\_noise\_sampler.p\_mean} & -1.2 \\
\texttt{train\_noise\_sampler.p\_std} & 1.5 \\
\texttt{sigma\_data} & 16.0 \\
\texttt{gamma0} & 0.8 \\
\texttt{gamma\_min} & 1.0 \\
\texttt{noise\_scale\_lambda} & 1.003 \\
\texttt{step\_scale\_eta} & 1.5 \\
\hline
\end{tabular}
}
\end{minipage}
\end{table}

\begin{table}[ht!]
\centering
\caption{Training stages for OpenDDE. Stage IV(a) inherits the checkpoint from Stage III, but freezes all major modules and trains only the confidence head. Stage IV(b) also inherits the checkpoint from Stage III and increases the crop size from 544 to 768. Note: The Teddymer dataset was introduced in~\cite{didi2026scaling}, and we reproduced the dataset for model training.}
\label{tab:training_stages}
\resizebox{\linewidth}{!}{
\begin{tabular}{lcccccc}
\toprule
\textbf{Parameter} & \textbf{Stage warmup} & \textbf{Stage I} & \textbf{Stage II} & \textbf{Stage III} & \textbf{Stage IV(a)} & \textbf{Stage IV(b)} \\
\midrule
Steps & 10k & 60k & 15k & 15k & 10k & 10k \\
GPUs & 50*8 80G & 50*8 80G & 50*8 80G & 50*8 80G & 50*8 80G & 8*8 141G \\
\hline
Crop tokens & 384 & 384 & 544 & 544 & 544 & 768 \\
Max atoms & 3,500 & 3,500 & 6,000 & 6,000 & 6,000 & 7,500 \\
Max MSA depth & 2,048 & 2,048 & 2,048 & 2,048 & 2,048 & 2,048 \\
MSA dropout probability & 0.2 & 0.2 & 0.2 & 0.2 & 0.2 & 0.2 \\
Diffusion batch size & 64 & 64 & 32 & 32 & 32 & 32 \\
Smooth LDDT loss & $\checkmark$ & $\checkmark$ & $\checkmark$ & $\checkmark$ & $\times$ & $\checkmark$ \\
Shape Complementary loss & $\checkmark$ & $\checkmark$ & $\checkmark$ & $\checkmark$ & $\times$ & $\checkmark$ \\
Train structure prediction & $\checkmark$ & $\checkmark$ & $\checkmark$ & $\checkmark$ & $\times$ & $\checkmark$ \\
Train \textit{de novo} design & $\times$ & $\times$ & $\times$ & $\checkmark$ & $\times$ & $\checkmark$ \\
\textit{De novo} design probability& 0 & 0 & 0 & 0.1 & 0 & 0.2 \\
Train confidence head & $\checkmark$ & $\checkmark$ & $\checkmark$ & $\checkmark$ & $\checkmark$ & $\checkmark$ \\
\hline
Weighted PDB~\cite{bytedance2025protenix}        & 70\% & 50\% & 35\% & 40\% & 50\% & 40\% \\
AFDB-multimer~\cite{han2026alphafold}       & 4\% & 5\% & 11\% & 9\% & 5\% & 9\% \\
Teddymer Reproduction      & 4\% & 5\% & 11\% & 9\% & 5\% & 9\% \\
MGnify long monomer~\cite{openfold3-preview}  & 10\% & 28\% & 30\% & 25\% & 20\% & 20\% \\
MGnify short monomer~\cite{openfold3-preview} & 4\% & 5\% & 4\% & 4\% & 4\% & 4\% \\
Swiss-Prot~\cite{varadi2022alphafold}           & 6\% & 5\% & 4\% & 4\% & 5\% & 4\% \\
Disordered~\cite{openfold3-preview}           & 2\% & 2\% & 2\% & 1\% & 1\% & 1\% \\
SAbDab~\cite{dunbar2014sabdab}              & 0\% & 0\% & 3\% & 8\% & 10\% & 13\% \\
\bottomrule
\end{tabular}
}
\end{table}

Compared with AlphaFold3, the main architectural difference is that \textsc{OpenDDE} scales the Pairformer hidden dimension from 128 to 384, which increases the Pairformer parameter count by approximately \(3\times\) and its computational cost by approximately \(9\times\). Beyond this scaling, \textsc{OpenDDE} introduces atomic-level reasoning blocks (Section~\ref{sec.Reasoning in Biomolecular Foundation Models}), shape-complementarity losses for interface fitting (Section~\ref{sec.Learning the Lock-and-Key Biomolecular Interactions}), and a unified training formulation for structure prediction and \textit{de novo} design (Section~\ref{sec.Structure Prediction and De Novo Design in One Framework}).

\subsection{Training Data}

Our training data follow the data construction protocols used in AlphaFold3~\cite{abramson2024accurate}, Protenix~\cite{bytedance2025protenix}, and ESMFold2~\cite{candido2026language}. To avoid test-set leakage, all ground-truth structures and template databases are filtered using a cutoff date of 2021-09, as we directly adopted from~\cite{abramson2024accurate, bytedance2025protenix, openfold3-preview, candido2026language}.

We collect training examples from multiple structure and sequence resources, including weighted PDB data from Protenix~\cite{bytedance2025protenix}, AFDB multimer data from Han et al.~\cite{han2026alphafold}, reproduced Teddymer dataset according to DiDi et al.~\cite{didi2026scaling}, MGnify long and short monomers reported in AlphaFold3~\cite{abramson2024accurate} and released by OpenFold3-Preview~\cite{openfold3-preview}, Swiss-Prot entries from AlphaFold DB~\cite{varadi2022alphafold}, disordered motifs reported in AlphaFold3~\cite{abramson2024accurate} and released by OpenFold3-Preview~\cite{openfold3-preview}, and SAbDab antibody structures reported in ESMFold2~\cite{candido2026language} and collected from SAbDab~\cite{dunbar2014sabdab}.

\subsection{Training Schedule}
\textsc{OpenDDE} is trained in four major stages with one additional warmup stage. For reproducibility, we report the training stages and the corresponding data distributions in Table~\ref{tab:training_stages}.

\textsc{OpenDDE} is initially trained on Ampere  GPUs with a warmup learning rate of 0.0012 for 10k steps, followed by a learning-rate decay factor of 0.95 every 60k steps. After the first three stages, we conduct two Stage IV variants: Stage IV(a), in which all major modules are frozen and only the confidence module is trained, and Stage IV(b), in which the full model is continuously trained on Hopper GPUs for larger crop size.

\textbf{Training Intuition} \quad 
The intuition behind the stage-wise data distribution is to train \textsc{OpenDDE} from precision, to breadth, and then back to precision. In the early stages, we emphasize large-scale experimentally validated structures so that the model first learns accurate folding patterns and reliable local geometry from experimental data. In the middle stages, we increase the use of distilled datasets to broaden coverage over sequence and structure space, allowing the model to learn more general fold-level and geometric priors. In the later stages, we shift the distribution back toward high-quality experimental and task-specific data to improve atomic accuracy, interface fitting, and confidence calibration. At the same time, the crop size is gradually increased, enabling the model to first learn stable local geometry and then adapt to longer-range interactions and larger biomolecular complexes.

\subsection{Training Computation}

We spent five weeks processing the metadata into trainable examples using \(10 \times 8\) NVIDIA 80GB GPUs. This preprocessing stage included data downloading, feature processing, and MSA search.

\textsc{OpenDDE} was then trained on \(50 \times 8\) NVIDIA 80GB Ampere GPUs for six weeks, followed by one additional week on \(8 \times 8\) NVIDIA 141GB Hopper GPUs. Additionally, we prepared \(5 \times 8\) NVIDIA 80GB Ampere GPUs to perform online validation. In total, \textsc{OpenDDE} required approximately 414K GPU-hours for training, that is 54 years in a single computing unit.

\section{OpenDDE-preview}

\begin{figure*}[ht!]
\centering

\begin{subfigure}[t]{1.0\textwidth}
    \centering
    \caption{Overall \textsc{OpenDDE} architecture.}
    \vspace{0.2cm}
    \includegraphics[width=\linewidth]{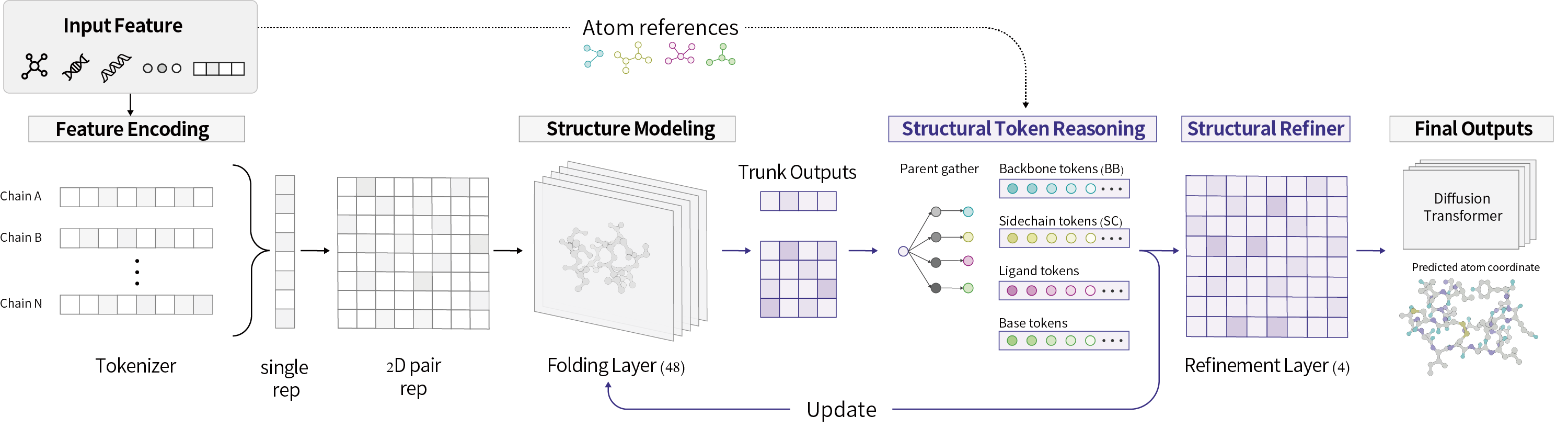}
    \label{fig:opendde_architecture}
\end{subfigure}

\begin{subfigure}[t]{0.64\textwidth}
    \centering
    \caption{Atom37-aware structural-token reasoning.}
    \vspace{0.4cm}
    \includegraphics[width=\linewidth]{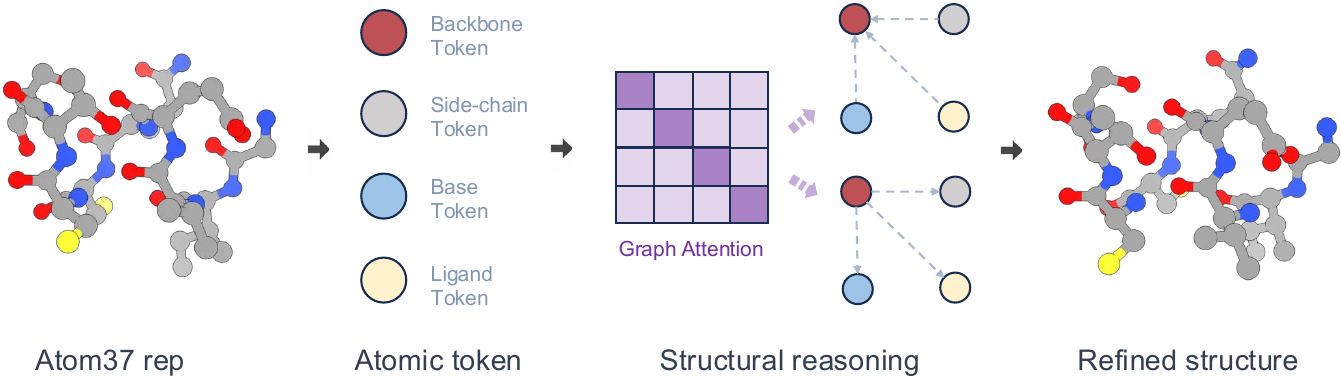}
    \label{fig:token_reason}
\end{subfigure}
\hfill
\begin{subfigure}[t]{0.32\textwidth}
    \centering
    \caption{Atomic shape complementary.}
    \vspace{0.3cm}
    \includegraphics[width=\linewidth]{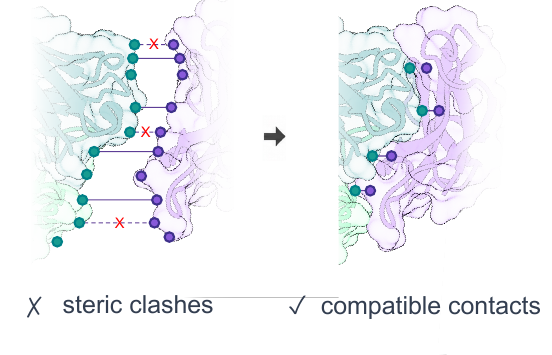}
    \label{fig:shape_comp}
\end{subfigure}


\caption{
Overview of \textsc{OpenDDE} and its unified structure-reasoning framework.
(a) \textsc{OpenDDE} builds residue-token and pair representations, expands them into structural tokens, and generates all-atom coordinates through diffusion.
(b) Atom37-level geometry is organized into semantic structural tokens, enabling reasoning over backbone, side-chain, base, and ligand roles before all-atom refinement.
(c) Shape-complementarity losses guide interface fitting by rewarding compatible contacts while penalizing gaps and steric clashes.
}
\label{fig:opendde_overview}
\end{figure*}

\textsc{OpenDDE} is a coarse-to-fine biomolecular structure generation model that unifies structure prediction and conditional design within one diffusion framework (Details can be found in Appendix~\ref{app:a}). Given sequence, atom, MSA, template, and constraint features, \textsc{OpenDDE} first builds residue-token representations with a Pairformer-style trunk. These residue-level states capture global relational context across chains, residues, ligands, and other molecular units. The model then expands each residue-token representation into chemically meaningful structural tokens, including protein backbone, side chain, nucleic-acid backbone, base, and ligand/atom tokens. A Structural Refiner performs fine-grained relational reasoning over these structural tokens before a diffusion-based coordinate generator denoises atom positions conditioned on the refined representations. Confidence and distance heads provide residue-level and structure-level prediction outputs.

\subsection{Reasoning in Biomolecular Foundation Models}

We define structure reasoning as latent relational inference over molecular tokens before coordinate generation. Rather than directly mapping sequence tokens to atom coordinates, \textsc{OpenDDE} first constructs single and pair representations that encode hypotheses about molecular geometry. The residue-token trunk captures global context, while the structural-token branch decomposes each residue or molecular unit into finer chemical components and refines their interactions.

This gives a coarse-to-fine reasoning path:
\[
(S^{r}, Z^{r})
\rightarrow
(S^{s}, Z^{s})
\rightarrow
X,
\]
where \((S^{r}, Z^{r})\) are residue-level single and pair states, \((S^{s}, Z^{s})\) are refined structural-token states, and \(X\) denotes all-atom coordinates. Pair-conditioned attention and triangular pair updates allow the model to reason not only about individual token-token relations, but also about geometric consistency across indirect paths. In this view, structural tokens form an intermediate layer between residue-level context and atom-level coordinates, enabling the model to represent backbone, side-chain, base, and ligand interactions before denoising atoms.

\subsection{Learning the \textit{Lock-and-Key} Biomolecular Interactions}
\label{sec.Learning the Lock-and-Key Biomolecular Interactions}

Accurate molecular structure generation requires more than placing atoms near their native positions. For complexes, the interacting surfaces must face each other, maintain realistic gaps, avoid steric clashes, and form locally compatible packing. To encourage this behavior, \textsc{OpenDDE} augments diffusion training with geometry-aware objectives that supervise both interface complementarity and all-atom refinement.

The shape-complementarity objective compares predicted and native interface geometry using token-level surface orientation, cross-chain spacing, and clash-aware contact quality. This guides the model toward binding interfaces that are physically compatible rather than merely close under coordinate distance. In parallel, local-frame and torsion supervision improve all-atom packing by constraining side chains, nucleic-acid bases, and ligand/atom tokens in chemically meaningful coordinate systems. Together, these losses teach the model the ``lock-and-key'' geometry of biomolecular interactions: global chains must fit, and local atoms must pack.

\subsection{Structure Prediction and \textit{De Novo} Design in One Framework}
\label{sec.Structure Prediction and De Novo Design in One Framework}

\textsc{OpenDDE} treats structure prediction and de novo design as two instances of the same conditional denoising problem. The input complex is represented as a partially observed molecular system, where some atoms can be fixed as structural context and the remaining atoms are generated as targets. Standard structure prediction is the special case in which no atoms are fixed, while conditional design fixes one or more chains, motifs, pockets, or molecular contexts.

This is implemented through known-target atom masks:
\[
m^{\mathrm{known}}_a \in \{0,1\},
\qquad
m^{\mathrm{target}}_a \in \{0,1\}.
\]
During conditional diffusion, known atoms remain fixed as structural context, while target atoms are noised and denoised:
\[
p_\theta(X^{\mathrm{target}} \mid \mathcal{I}, X^{\mathrm{known}}).
\]
When \(m^{\mathrm{known}}=0\), the same model performs full structure prediction. When \(m^{\mathrm{known}}\neq0\), it performs conditional generation or design around the provided structure.

This unified formulation allows prediction and design to share the same representation learning pipeline: residue-token reasoning, structural-token refinement, and all-atom diffusion. As a result, \textsc{OpenDDE} does not require a separate design-specific architecture. Instead, de novo design becomes masked molecular generation under structural conditions, while structure prediction remains the fully unconditioned case.

\section{Next Steps}

We are actively collaborating with the NVIDIA Fold-CP team~\cite{lin2026fold} to improve the computational scalability of OpenDDE. Our immediate goal is to enable efficient parallelization of a single biomolecular sample across multiple GPUs, while also improving GPU utilization by packing multiple smaller samples onto the same GPU when memory allows.

Building on these system-level improvements, we plan to scale \textsc{OpenDDE} beyond one billion parameters and train the next-generation model on NVIDIA B300 GPUs. This will allow us to further test model and data scaling laws for biomolecular foundation models and to extend \textsc{OpenDDE} toward larger complexes, higher-resolution all-atom refinement, and more challenging structure design tasks.

\section{Limitations}

Despite its strong empirical performance, \textsc{OpenDDE} has several limitations. First, our comparison with IsoDDE is necessarily constrained by the limited public information available about the system. We do not have access to its full training recipe, data mixture, inference procedure, post-training strategy, or engineering optimizations. As a result, although \textsc{OpenDDE} reaches the level of accuracy reported for IsoDDE on several co-folding benchmarks, we cannot yet determine which differences arise from model architecture, data processing, scaling, distillation, or inference-time procedures. Further work is needed to systematically dissect these factors and to understand which components are essential for robust all-atom co-folding.

Second, this work focuses primarily on protein-protein and antibody-antigen interactions rather than protein-small molecule complexes. This choice reflects both our scientific objective and our empirical observations. \textsc{OpenDDE} is designed around biomolecular co-folding as a structural core for modelling macromolecular interfaces, and our current training and evaluation emphasis follows this setting. In preliminary analyses, we observed limited correlation between performance on protein-protein interfaces and protein-ligand pose prediction, suggesting that these regimes may depend on partly distinct data, inductive biases, and evaluation criteria. We therefore do not claim that the present model is optimized for ligand docking, virtual screening, or affinity ranking. Extending \textsc{OpenDDE} to these tasks will likely require ligand-focused data curation, chemistry-aware post-training, and task-specific evaluation.

Third, \textsc{OpenDDE} should be viewed as the folding-centered foundation of an extensible drug-discovery engine, not as a complete drug-discovery system. The present release establishes the structural modelling core and demonstrates that open co-folding models can scale competitively with closed systems. However, downstream capabilities such as molecular design, affinity prediction, conformational ensemble modelling, active learning, and experimental feedback remain future extensions rather than fully validated components of the current system. Similarly, our scaling analyses identify promising directions within the model and data regimes explored here, but they do not exhaust the possible scaling behavior of all-atom biomolecular models. Continued community-driven development, broader benchmarks, and more systematic ablations will be required to turn \textsc{OpenDDE} from a strong open folding model into a mature open drug-discovery engine.

\section*{Acknowledgements}

We thank the AlphaFold, Protenix, and OpenFold teams for releasing their codebases and, in some cases, providing training data resources. We also thank the NVIDIA infrastructure and architecture teams for developing computing kernels that accelerate model training and inference.

\section*{Contributions}
Hua, Ma for architecture;\
Hua, Ma for training;\
Hua, Ma for inference;\
Ma for data;\
Lu, Tao for 2026ARK-AB;\
Wang for parallel computing;\
Hua, Ma, Zheng, Wang, Rehman for writing;\
Hua, Zheng for supervision.
Zheng for organization and conceptualization.

\bibliographystyle{unsrtnat}
\bibliography{references}

\appendix 

\section{Model Architecture}
\label{app:a}
\textsc{OpenDDE} is organized as a coarse-to-fine molecular structure generation model. 
It first encodes input sequence, atom, MSA, template, and constraint features into 
a residue-token representation. A residue-level trunk, built on Pairformer-style 
single and pair updates, performs global relational reasoning over residues and 
molecular units. The resulting residue-token outputs are then expanded into 
structural tokens, separating chemically meaningful components such as protein 
backbone, side chain, nucleic-acid backbone, base, and ligand/atom tokens. A 
Structural Refiner further updates these structural-token representations to 
model fine-grained geometric interactions. Finally, a diffusion-based coordinate 
generator denoises atom positions conditioned on the refined structural-token 
features, while confidence and distance heads produce residue-level or 
structure-level prediction outputs.

\subsection{Reasoning in Biomolecular Foundation Models}
\label{sec.Reasoning in Biomolecular Foundation Models}
In this work, structure reasoning denotes neural relational inference over molecular tokens: the model builds, conditions, and refines latent pairwise representations before using them as geometric constraints for atom-level denoising.

We refer to this process as structure reasoning because the model does not directly map sequence tokens to coordinates. Instead, it first constructs and updates latent relational states that encode hypotheses about molecular geometry. The pair representation \(Z\) stores token-token structural relations, while the single representation \(S\) stores token-wise contextual states. Through structural-token expansion and refinement, \textsc{OpenDDE} infers how residue-level context should be decomposed into backbone, side-chain, base, and ligand/atom units, and how these units should interact before coordinate generation.

\subsubsection{Structure Reasoning over Structural Tokens}
\textsc{OpenDDE} performs structure reasoning by converting residue-level trunk outputs into a structural-token graph and then applying relational message passing over this graph. Each structural token is treated as a node, while the pair representation between two structural tokens is treated as an edge:
\[
\mathcal{G}^{s} = (\mathcal{V}^{s}, \mathcal{E}^{s}),
\qquad
u \in \mathcal{V}^{s}, \quad
z_{uv}^{s} \in \mathcal{E}^{s}.
\]

The Structural Token Expander initializes this graph from the residue-token trunk:
\[
s_u^{0}
=
s_{\pi(u)}^{r}
+
\mathrm{MLP}(s_{\pi(u)}^{r})
+
e_{\rho(u)},
\]
\[
z_{uv}^{0}
=
z_{\pi(u),\pi(v)}^{r}
+
P_{\rho(u),\rho(v)}(z_{\pi(u),\pi(v)}^{r})
+
b(\phi_{uv}),
\]
where \(\pi(u)\) is the parent residue of structural token \(u\), \(\rho(u)\) is its structural role, and \(\phi_{uv}\) encodes explicit structural relations such as same-residue membership, backbone--sidechain/base pairing, and polymer backbone adjacency.

The Structural Refiner then performs reasoning by repeatedly updating node and edge states. In each layer, structural tokens attend to other structural tokens under pair-conditioned attention:
\[
\alpha_{uv}^{(l)}
=
\mathrm{softmax}_{v}
\left(
\frac{q_u^{(l)} {k_v^{(l)}}^{\top}}{\sqrt{d}}
+
g(z_{uv}^{(l)})
+
a_{uv}^{s}
\right),
\]
\[
s_u^{(l+1)}
=
s_u^{(l)}
+
\sum_v \alpha_{uv}^{(l)} V_v^{(l)}.
\]

The pair representation is also refined through triangular updates:
\[
z_{uv}^{(l+1)}
=
z_{uv}^{(l)}
+
\mathrm{TriUpdate}
\left(
z_{uv}^{(l)}, \{z_{uk}^{(l)}, z_{kv}^{(l)}\}_{k}
\right).
\]

This triangular update is important because it allows the model to reason about geometric consistency through indirect paths. For example, the relation between two tokens \(u\) and \(v\) can be updated using their shared relations with a third token \(k\). Thus, the model does not reason about each pair independently; it enforces consistency across many token-token-token configurations.

Therefore, structure reasoning in \textsc{OpenDDE} can be summarized as:
\[
(S^{r}, Z^{r})
\rightarrow
(S^{s,0}, Z^{s,0})
\rightarrow
(S^{s,L}, Z^{s,L})
\rightarrow
X.
\]

Here, \(S^{s,L}\) and \(Z^{s,L}\) are refined structural-token representations that encode both global residue-level context and local chemical geometry. These representations serve as geometric constraints for the downstream diffusion module, which generates atom-level coordinates.

\subsubsection{Intuition Behind Structure Reasoning}
The intuition behind structure reasoning in \textsc{OpenDDE} is that molecular structure should not be generated as an immediate coordinate regression problem. Instead, the model first builds an internal relational representation of the molecule, where tokens encode structural units and pair features encode their geometric and chemical dependencies.

At the residue-token level, the trunk captures global context from sequence, MSA, templates, chain identity, and pairwise priors. This stage answers a coarse structural question: which residues or molecular units are likely to be close, constrained, or functionally coupled? However, residue tokens are too coarse to fully describe local geometry, since a single residue contains chemically distinct components such as backbone atoms, side-chain atoms, or nucleic-acid bases.

Structural token expansion introduces a finer representation that exposes these chemically meaningful sub-units. A residue is no longer treated as a single object; instead, it is decomposed into structural roles such as backbone, side chain, base, or ligand/atom tokens. This allows the model to reason over the parts that directly determine atomic geometry.

The Structural Refiner then performs relational reasoning over this expanded structural-token graph. Tokens exchange information through pair-conditioned attention and triangular pair updates, allowing local structural components to adjust to one another. For example, a side chain is conditioned not only on its parent residue but also on neighboring backbones, nearby side chains, ligand atoms, and longer-range pair constraints. Thus, the model refines a globally consistent structural hypothesis before atom-level coordinate generation.

In this view, structure reasoning is a coarse-to-fine process:
\[
\text{sequence/context}
\rightarrow
\text{residue-level relations}
\rightarrow
\text{structural-token relations}
\rightarrow
\text{atom-level coordinates}.
\]

The key benefit is that \textsc{OpenDDE} separates the problem of understanding molecular relationships from the problem of generating coordinates. The trunk and refiner infer geometric constraints in latent space, while the diffusion module uses these refined constraints to produce physically plausible atomic structures.

\subsubsection{Between Atom37 Representation and Structure Reasoning}
The Atom37 representation describes each protein residue using a fixed set of up to 37 atom slots:
\[
X_i^{37} = \{x_{i,a}\}_{a=1}^{37},
\qquad
m_{i,a} \in \{0,1\},
\]
where \(x_{i,a}\) is the coordinate of atom slot \(a\) in residue \(i\), and \(m_{i,a}\) indicates whether this atom exists for the corresponding amino-acid type. Thus, Atom37 provides a residue-local atom layout: it tells the model which atoms should be generated for each residue and how atom identities are consistently indexed across residues.

However, Atom37 itself does not perform structure reasoning. It is mainly an atom-level coordinate representation. The reasoning happens in the latent token representations before coordinates are produced. At the residue level, \textsc{OpenDDE} first learns contextual single and pair representations:
\[
S^r = \{s_i^r\}_{i=1}^{N_r},
\qquad
Z^r = \{z_{ij}^r\}_{i,j=1}^{N_r},
\]
where \(S^r\) captures residue-wise context and \(Z^r\) captures residue-residue geometric relations.

Structure reasoning can be viewed as the process of using these latent representations to infer constraints that will later determine Atom37 coordinates:
\[
(S^r, Z^r)
\rightarrow
\text{geometric constraints}
\rightarrow
X^{37}.
\]

In OpenDDE, this idea is further refined by introducing structural tokens. Instead of mapping one residue token directly to all Atom37 atom slots, the model decomposes a residue into chemically meaningful structural units, such as backbone and side chain:
\[
i
\rightarrow
\{u: \pi(u)=i\},
\]
where \(u\) is a structural token and \(\pi(u)\) is its parent residue.

Under this view, Atom37 is the final atom-level coordinate container, while structure reasoning is the latent process that decides how those atom slots should be spatially arranged. The structural-token branch acts as an intermediate layer between residue-level reasoning and Atom37-like atom-level geometry:
\[
\text{residue tokens}
\rightarrow
\text{structural tokens}
\rightarrow
\text{atom coordinates}.
\]

Therefore, Atom37 and structure reasoning are complementary. Atom37 defines the atom-level output space, whereas structure reasoning provides the learned geometric constraints that make the generated atom coordinates globally and locally consistent.

\subsection{Learning the \textit{Lock-and-Key} Biomolecular Interactions}

\textsc{OpenDDE} augments diffusion-based coordinate generation with geometry-aware
auxiliary objectives. After residue-token reasoning and structural-token
refinement, the model predicts all-atom coordinates
\[
X = \{x_a \in \mathbb{R}^3\}_{a=1}^{N_{\mathrm{atom}}},
\]
conditioned on structural-token representations \((S^s, Z^s)\). The predicted
coordinates are then supervised not only by point-wise coordinate errors, but
also by losses that describe interface fitting and local atomic packing.

\subsubsection{Shape-Complementarity Losses Guide Interface Fitting}

For molecular complexes, accurate fitting requires more than placing chains
near each other. The contacting surfaces should face each other, maintain an
appropriate gap, avoid steric clashes, and exhibit complementary surface
normals. \textsc{OpenDDE} introduces a differentiable shape-complementarity objective to
encode these geometric requirements.

Given predicted coordinates \(X\), each active token \(u\) is assigned a center
\(c_u\). In residue-token space, this center is usually selected from a
representative atom. In structural-token space, side-chain tokens can use the
centroid of supervised atoms, allowing the loss to better describe side-chain
surface geometry.

A local surface normal is estimated from the same-chain atomic density around
the token center:
\[
g_u =
\sum_{a \in \mathcal{C}(u)}
\exp\left(
-\frac{\|c_u - x_a\|^2}{2\sigma^2}
\right)
\frac{c_u - x_a}{\sigma^2},
\qquad
n_u = \frac{g_u}{\|g_u\|+\epsilon},
\]
where \(\mathcal{C}(u)\) denotes atoms from the same chain as token \(u\).
This gradient points away from the local molecular density and approximates a
surface normal.

For a cross-chain token pair \((u,v)\), let
\[
d_{uv} = c_v - c_u, \qquad
\hat{d}_{uv} = \frac{d_{uv}}{\|d_{uv}\|+\epsilon}.
\]
\textsc{OpenDDE} computes several geometric factors:
\[
f_{\mathrm{face}}(u,v)
=
\mathrm{ReLU}(n_u^\top \hat{d}_{uv})
\,
\mathrm{ReLU}(n_v^\top (-\hat{d}_{uv})),
\]
\[
f_{\mathrm{opp}}(u,v)
=
\frac{1 - n_u^\top n_v}{2},
\]
\[
f_{\mathrm{gap}}(u,v)
=
\exp\left(
-\left(
\frac{\|d_{uv}\|-\mu_{\mathrm{gap}}}{\tau_{\mathrm{gap}}}
\right)^2
\right),
\]
\[
f_{\mathrm{clash}}(u,v)
=
1 -
\sigma_{\mathrm{sigmoid}}
\left(
\frac{d_{\mathrm{clash}}-\|d_{uv}\|}{\tau_{\mathrm{clash}}}
\right).
\]

The pairwise shape-complementarity score is then defined as
\[
q_{uv}
=
f_{\mathrm{face}}(u,v)
\,
f_{\mathrm{opp}}(u,v)
\,
f_{\mathrm{gap}}(u,v)
\,
f_{\mathrm{clash}}(u,v),
\]
computed only for valid cross-chain token pairs within an interface cutoff.
Token-level and global scores are obtained by pooling these pair scores:
\[
q_u = \mathrm{Pool}_{v}(q_{uv}),
\qquad
q_{\mathrm{global}} = \mathrm{Mean}_{u}(q_u).
\]

During training, the same fields are computed from the ground-truth coordinates,
yielding \(q^{*}_{uv}\), \(q^{*}_u\), and \(q^{*}_{\mathrm{global}}\). The
shape-complementarity objective is:
\[
\mathcal{L}_{\mathrm{shape}}
=
\lambda_p
\mathrm{Huber}(q_{uv}, q^{*}_{uv})
+
\lambda_t
\mathrm{Huber}(q_u, q^{*}_u)
+
\lambda_g
\mathrm{Huber}(q_{\mathrm{global}}, q^{*}_{\mathrm{global}}).
\]

This loss guides the model toward interface geometries that are not only close
in Euclidean distance but also physically compatible in surface orientation,
spacing, and anti-clash behavior. It is especially useful for protein-protein,
protein-ligand, and protein-nucleic-acid fitting, where correct binding geometry
depends on surface complementarity rather than isolated atom placement.

\subsubsection{All-Atom Refinement Improves Local Packing}

\textsc{OpenDDE} further improves local packing by supervising coordinates in
all-atom and local-frame spaces. The diffusion module predicts atom coordinates
conditioned on refined structural-token features:
\[
\hat{X}_0 =
D_{\theta}(X_t, t \mid S^s, Z^s, \mathcal{A}),
\]
where \(\mathcal{A}\) contains atom-to-token and atom-to-structural-token
mappings. This design allows each atom to receive both global pair context and
local chemical-role information.

A standard global coordinate loss can penalize atom displacement, but it may not
fully capture residue-internal geometry. Therefore, \textsc{OpenDDE} introduces
local-frame supervision. For each residue or structural token with a valid frame
\(T_u\), predicted and ground-truth atoms are expressed in the same local frame:
\[
\tilde{x}^{\mathrm{pred}}_a = T_u^{-1}\hat{x}_a,
\qquad
\tilde{x}^{\mathrm{true}}_a = T_u^{-1}x^{*}_a.
\]
The local coordinate loss is:
\[
\mathcal{L}_{\mathrm{local}}
=
\frac{1}{|\mathcal{A}_u|}
\sum_{a \in \mathcal{A}_u}
\mathrm{SmoothL1}
\left(
\tilde{x}^{\mathrm{pred}}_a,
\tilde{x}^{\mathrm{true}}_a
\right).
\]

For protein residues, \textsc{OpenDDE} also supervises side-chain torsions:
\[
\mathcal{L}_{\chi}
=
1 - \cos(\hat{\chi} - \chi^{*}),
\]
which directly encourages correct rotameric geometry. For structural tokens,
side-chain and nucleic-acid base atoms are supervised in frames inherited from
their parent backbone token:
\[
\mathcal{L}_{\mathrm{sc/base}}
=
\sum_{u \in \mathcal{V}_{\mathrm{sc/base}}}
w_{\rho(u)}
\frac{1}{|\mathcal{A}_u|}
\sum_{a \in \mathcal{A}_u}
\mathrm{SmoothL1}
\left(
T_u^{-1}\hat{x}_a,
T_u^{-1}x^{*}_a
\right).
\]

The full geometry objective can be summarized as:
\[
\mathcal{L}_{\mathrm{geom}}
=
\mathcal{L}_{\mathrm{diff}}
+
\lambda_{\mathrm{shape}}\mathcal{L}_{\mathrm{shape}}
+
\lambda_{\mathrm{local}}\mathcal{L}_{\mathrm{local}}
+
\lambda_{\mathrm{sc/base}}\mathcal{L}_{\mathrm{sc/base}}
+
\lambda_{\chi}\mathcal{L}_{\chi}.
\]

This all-atom refinement pathway improves local packing because it supervises
atoms in chemically meaningful coordinate systems. Backbone, side-chain, base,
and ligand/atom tokens are not treated as anonymous points; they are connected
through structural roles, parent-residue frames, atom mappings, and pairwise
geometric context. As a result, the model learns to place side chains, bases,
and interface atoms in locally consistent configurations before producing the
final all-atom structure.

\subsection{Structure Prediction and \textit{De Novo} Design in One Framework}

\textsc{OpenDDE} formulates structure prediction and de novo design as two instances of
the same conditional denoising problem. The key idea is to represent the input
complex as a partially observed molecular system, where some atoms may be treated
as fixed structural context and the remaining atoms are treated as prediction or
design targets.

For each training example, we define two atom masks:
\[
m^{\mathrm{known}}_a \in \{0,1\},
\qquad
m^{\mathrm{target}}_a \in \{0,1\},
\]
where \(m^{\mathrm{known}}_a=1\) indicates atoms provided as structural
condition, and \(m^{\mathrm{target}}_a=1\) indicates atoms to be generated or
predicted. In standard structure prediction, no chain is fixed:
\[
m^{\mathrm{known}} = 0,
\qquad
m^{\mathrm{target}} = m^{\mathrm{resolved}}.
\]
In conditional design, one or more chains, motifs, pockets, or structural
contexts are fixed:
\[
m^{\mathrm{known}} \neq 0,
\qquad
m^{\mathrm{target}} = m^{\mathrm{resolved}} - m^{\mathrm{known}}.
\]

Thus, the same model can be trained on both tasks by changing only the data
condition masks, rather than introducing a separate architecture.

\subsubsection{Known-Chain Data Conditioning}

During training, \textsc{OpenDDE} samples partial-complex conditioning cases from
multi-chain structures. A polymer chain is selected as known structural context,
while the remaining resolved atoms are treated as target atoms. This produces
the conditioning fields
\[
\mathcal{C}
=
\{
m^{\mathrm{known}},
m^{\mathrm{target}},
\mathrm{is\_conditioned}
\}.
\]

The input feature dictionary therefore contains both ordinary structure
prediction features and conditional-generation features:
\[
\mathcal{I}
=
\{
\mathrm{sequence},
\mathrm{MSA},
\mathrm{template},
\mathrm{atom\ features},
\mathrm{token\ features},
\mathcal{C}
\}.
\]

The residue-token trunk and structural-token branch encode the full molecular
system:
\[
\mathcal{I}
\rightarrow
(S^r, Z^r)
\rightarrow
(S^s, Z^s),
\]
where \((S^r, Z^r)\) are residue-token representations and
\((S^s, Z^s)\) are structural-token representations. The condition masks do not
require a separate design model; they only change how noise, alignment, and loss
are applied.

\subsubsection{Conditional Diffusion Objective}

Let \(X_0\) be the ground-truth all-atom coordinates. \textsc{OpenDDE} first applies a
shared random rigid augmentation:
\[
\bar{X}_0 = R X_0 + t.
\]
For ordinary structure prediction, Gaussian noise is added to all supervised
atoms:
\[
X_t = \bar{X}_0 + \sigma_t \epsilon.
\]

For data-conditioned design, noise is added only to target atoms, while known
atoms remain fixed as structural context:
\[
X_t
=
m^{\mathrm{known}} \odot \bar{X}_0
+
m^{\mathrm{target}} \odot
(\bar{X}_0 + \sigma_t \epsilon).
\]

The denoising network predicts coordinates conditioned on both the latent
representations and the partially observed structure:
\[
\hat{X}_0
=
D_\theta
\left(
X_t, t
\mid
S^s, Z^s, \mathcal{I}, m^{\mathrm{known}}
\right).
\]

The training loss is then applied primarily on the target atoms:
\[
\mathcal{L}_{\mathrm{diff}}
=
\mathbb{E}_{t,\epsilon}
\left[
\left\|
m^{\mathrm{target}}
\odot
\left(
\hat{X}_0 - \bar{X}_0
\right)
\right\|^2
\right].
\]

This objective naturally supports both tasks. When
\(m^{\mathrm{known}}=0\), the model learns standard structure prediction. When
\(m^{\mathrm{known}}\neq0\), the model learns to generate missing or designable
parts conditioned on a fixed molecular context.

\subsubsection{Unified Prediction-Design Framework}

The unified framework can be summarized as:
\[
\text{full complex data}
\rightarrow
\text{condition sampling}
\rightarrow
(m^{\mathrm{known}}, m^{\mathrm{target}})
\rightarrow
\text{conditional diffusion training}.
\]

For structure prediction, the model receives sequence and feature information
and predicts the full structure:
\[
p_\theta(X \mid \mathcal{I}).
\]

For de novo design, the model receives an additional structural condition and
generates the target region:
\[
p_\theta
\left(
X^{\mathrm{target}}
\mid
\mathcal{I},
X^{\mathrm{known}}
\right).
\]

Both distributions are implemented by the same denoising model:
\[
D_\theta(X_t,t\mid S^s,Z^s,\mathcal{I},m^{\mathrm{known}}).
\]

This design has three advantages. First, prediction and design share the same
representation learning pipeline, including residue-token reasoning,
structural-token refinement, and all-atom diffusion. Second, the model learns
from ordinary structure data without requiring a separate curated design
dataset. Third, conditional design becomes a masked generation problem: fixed
chains, motifs, or binding contexts are provided as data conditions, while the
model denoises and generates the target molecular components around them.

In this sense, \textsc{OpenDDE} does not treat structure prediction and de novo design as
separate tasks. Instead, both are unified as conditional structure generation
under different choices of the known-target partition.

\section{Experiment}

\subsection{Overall Performance}

\begin{figure}[t!]
\centering
    \includegraphics[width=\linewidth]{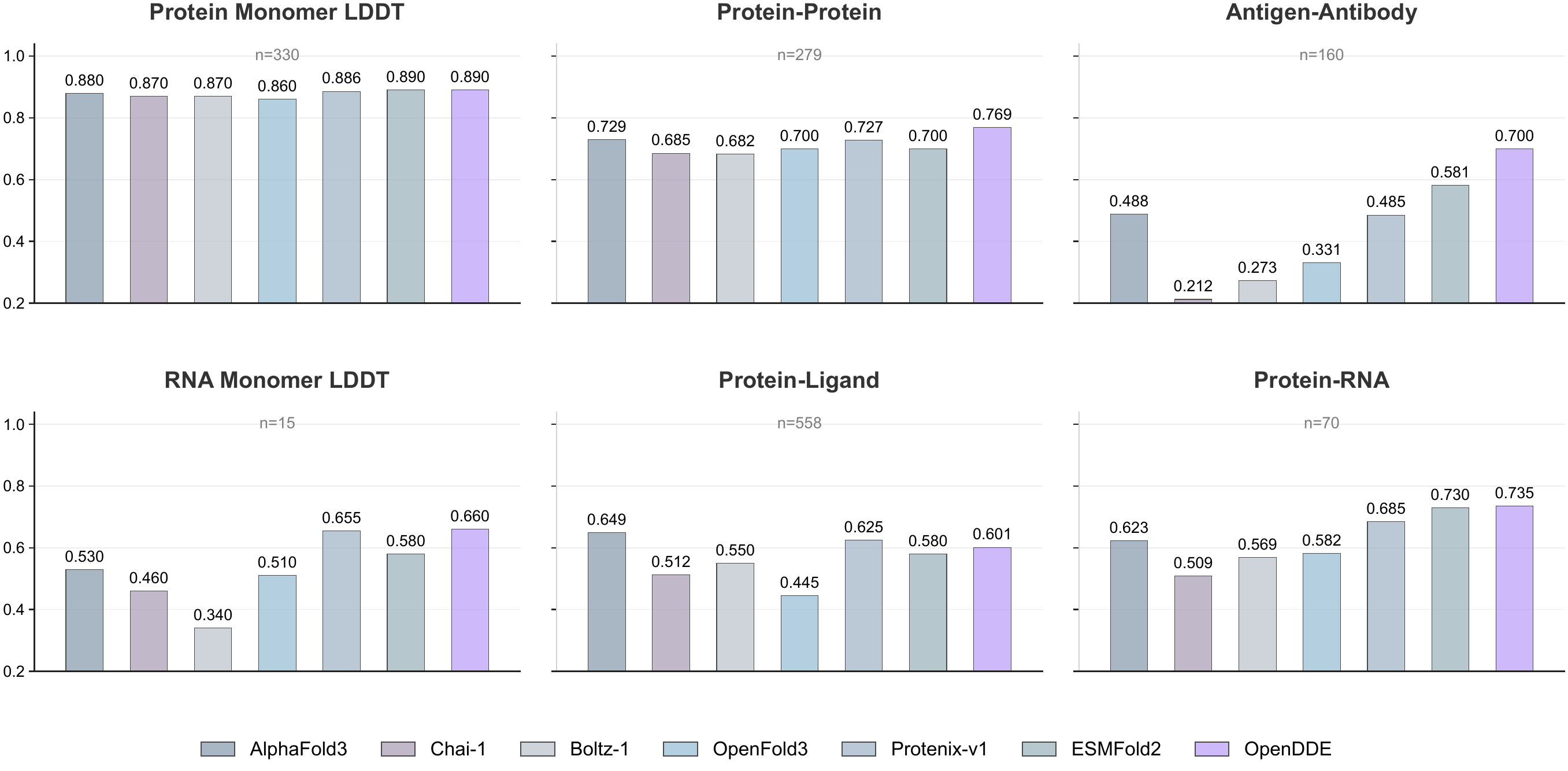}
    \caption{\textbf{Structure prediction performance on FoldBench.} In the two monomer panels, the metric is mean LDDT. In the protein-protein, antigen-antibody, and protein-RNA panels, the metric is DockQ success rate. In the protein-ligand panel, the metric is ligand pose success rate under the corresponding FoldBench criterion.}
\label{fig:overall_performance}
\end{figure}

Overall, \textsc{OpenDDE} achieves strong and balanced performance across the FoldBench tasks (Figure~\ref{fig:overall_performance}), with particularly large gains on interaction-heavy systems. It reaches the best result on protein monomer LDDT, reaching 0.890, and obtains the highest protein-protein score at 0.769. The largest improvement appears on antigen-antibody complexes, where \textsc{OpenDDE} reaches 0.700, substantially outperforming ESMFold2 (0.581), AlphaFold3 (0.488), and Protenix-v1 (0.485). \textsc{OpenDDE} also remains competitive on RNA monomers (0.660), protein-ligand complexes (0.601), and protein-RNA complexes (0.735), where it outperforms or is close to the strongest baseline. These results suggest that \textsc{OpenDDE} preserves high single-chain structure accuracy while improving the more challenging cross-molecular reasoning needed for protein-protein and antibody-antigen interfaces.

\subsection{Test-Time Scaling}

\begin{figure}[ht!]
\centering
    \includegraphics[width=\linewidth]{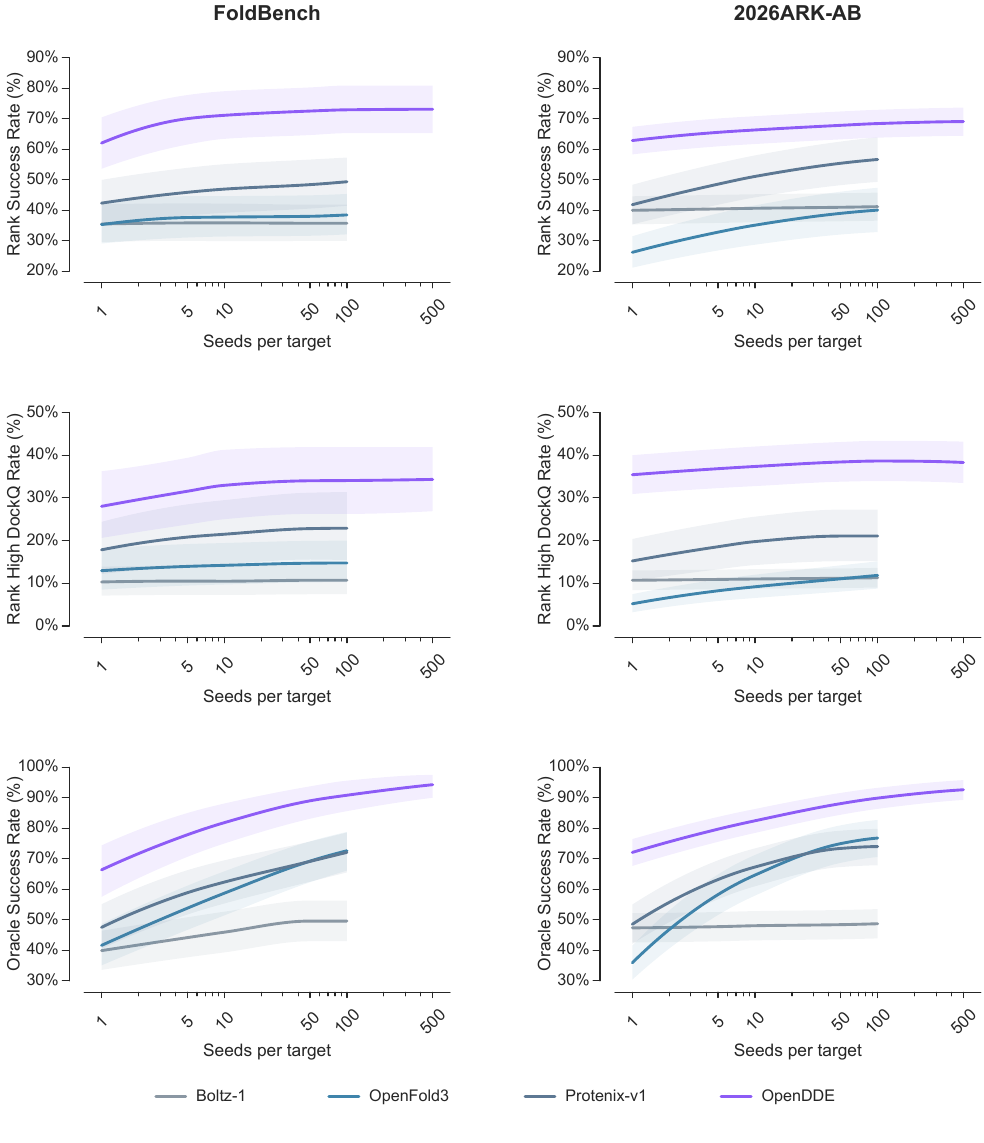}
    \vspace{-0.8cm}
    \caption{\textbf{Test-time scaling on FoldBench-AB and 2026ARK-AB.} \textit{Top:} Ranking-based selection. \textit{Middle:} Ranking-based selection of DockQ $>$ 0.8. \textit{Bottom}: Oracle-based selection.}
\label{fig:tts.appendix}
\end{figure}

We evaluate whether additional inference-time sampling improves antibody-antigen prediction performance (Figure~\ref{fig:tts.appendix}). For each target, we generate multiple stochastic samples with different random seeds and select the final prediction using the model ranking score. We report three settings: ranking-based success rate, ranking-based high-quality success rate with DockQ $>$ 0.8, and oracle success rate, where the best candidate among all generated samples is selected using the ground-truth DockQ. The oracle setting measures the maximal performance available in the sampled candidate set and separates sampling capacity from ranking quality.

As shown in Figure~\ref{fig:tts.appendix}, \textsc{OpenDDE} exhibits consistent test-time scaling on both FoldBench-AB and 2026ARK-AB. On FoldBench-AB, increasing the number of seeds improves the ranked success rate from roughly 62\% with one seed to about 73\% at large sampling budgets, while the ranked high-DockQ rate increases from about 28\% to 34\%. Similar behavior is observed on 2026ARK-AB, where \textsc{OpenDDE} improves from approximately 63\% to 69\% ranked success and maintains the strongest high-DockQ performance across seed counts. These gains indicate that additional samples expose better antibody-antigen conformations rather than simply producing redundant predictions.

The oracle curves show a larger scaling effect. On FoldBench-AB, \textsc{OpenDDE}'s oracle success rises from about 66\% with one seed to above 90\% with hundreds of seeds. On 2026ARK-AB, the oracle success similarly increases from about 67\% to nearly 90\%. This gap between ranked and oracle performance suggests that \textsc{OpenDDE} often generates high-quality antibody-antigen structures that are not always selected as the top-ranked prediction. Therefore, test-time scaling reveals strong latent sampling capacity, and further improvements in confidence calibration or candidate ranking could convert more of this oracle performance into realized prediction accuracy.

Across all three metrics, \textsc{OpenDDE} scales more favorably with inference-time compute than the compared open models, including Boltz-1, OpenFold3, and Protenix-v1. These results show that \textsc{OpenDDE} benefits predictably from larger sampling budgets and that its diffusion-based generation process provides a practical route for improving antibody-antigen prediction accuracy at test time.

\section{\textsc{OpenDDE} Parallel Computing}

\textsc{OpenDDE} integrates Fold-CP~\cite{lin2026fold} to enable memory-efficient inference for large biomolecular systems. The dominant memory bottleneck in \textsc{OpenDDE} arises from token-pair representations, whose size scales quadratically with the number of tokens. Given a pair representation
\[
Z \in \mathbb{R}^{N \times N \times C},
\]
Fold-CP partitions the two token dimensions across a two-dimensional context-parallel mesh. For a mesh rank \((p,q)\), the local pair block is
\[
Z_{pq} = Z[\mathcal{I}_p, \mathcal{I}_q, :],
\]
where \(\mathcal{I}_p\) and \(\mathcal{I}_q\) denote the token ranges assigned to the row and column dimensions of that rank. This decomposition avoids replicating the full \(N \times N\) pair tensor on every GPU, reducing per-device memory while preserving the original model computation.

OpenDDE applies this context-parallel layout to modules dominated by token-pair operations, including Pairformer-style trunk blocks, structural-token refinement, diffusion conditioning, and confidence prediction. Pair transitions, triangular updates, and pair-biased attention are evaluated on local pair blocks, with the required communication performed along the row and column axes of the context-parallel mesh. Thus, each device stores and updates only a shard of the pair representation, while collective communication reconstructs the information needed for operations that depend on shared row-wise, column-wise, or triangular context.

For attention with pair bias, the Fold-CP path follows blockwise exact attention computation, closely related to online softmax and memory-efficient attention. Rather than materializing the full attention score matrix, \textsc{OpenDDE} processes key, value, and pair-bias blocks incrementally across context-parallel ranks. The softmax statistics are accumulated block by block, so the resulting attention output is equivalent to standard softmax attention, while peak memory is reduced. This is particularly important for biomolecular systems because pair-derived attention biases scale with the same quadratic token-pair dimension.

The Fold-CP integration is an execution strategy rather than a different model. It does not change model weights, architecture, or inference semantics. Instead, it distributes the same \textsc{OpenDDE} computation across multiple GPUs, enabling larger complexes and longer token contexts to be processed under practical memory limits.

\begin{table}[t]
\centering
\caption{Runtime and memory comparison between single-GPU inference and CP4 inference.}
\label{tab:foldcp_runtime_memory}
\resizebox{\linewidth}{!}{
\begin{tabular}{llrrrr}
\toprule
\textbf{Input Length $N$}
& \textbf{Mode}
& \textbf{Forward Time (s)}
& \textbf{End-to-End Time (s)}
& \textbf{Peak GPU Memory (MiB)}
& \textbf{Torch Stage Peak Memory (MiB)} \\
\midrule
100  & single & 14.0   & 161.2  & 4{,}959  & 3{,}525  \\
100  & CP4    & 400.8  & 557.4  & 15{,}549 & 3{,}493  \\
250  & single & 36.2   & 180.8  & 10{,}619 & 9{,}381  \\
250  & CP4    & 393.7  & 543.7  & 17{,}353 & 6{,}557  \\
500  & single & 184.4  & 330.8  & 50{,}971 & 47{,}691 \\
500  & CP4    & 462.6  & 612.4  & 30{,}279 & 16{,}301 \\
900  & single & 468.4  & 615.2  & 56{,}855 & 52{,}991 \\
900  & CP4    & 766.4  & 920.2  & 56{,}443 & 28{,}660 \\
1200 & single & 982.2  & 1133.8 & 67{,}105 & 60{,}008 \\
1200 & CP4    & 969.9  & 1127.3 & 49{,}273 & 22{,}139 \\
1400 & single & 1636.7 & 1784.4 & 67{,}665 & 65{,}150 \\
1400 & CP4    & 1259.2 & 1398.3 & 59{,}237 & 28{,}924 \\
2000 & single & \multicolumn{4}{c}{OOM} \\
2000 & CP4    & 2568.2 & 2714.1 & 80{,}899 & 55{,}836 \\
\bottomrule
\end{tabular}
}
\vspace{0.5em}
\end{table}

Forward Time measures the model forward pass only.
End-to-End Time measures the full wall-clock runtime, including setup, input processing, model execution, post-processing, and output writing.
Peak GPU Memory is the maximum physical GPU memory observed by external monitoring.
Torch Stage Peak Memory is the peak memory recorded by PyTorch stage-level instrumentation.

\end{document}